\documentclass[10pt,twocolumn,letterpaper]{article}

\usepackage{cvpr}
\usepackage{times}
\usepackage{epsfig}
\usepackage{graphicx}
\usepackage{amsmath}
\usepackage{mathrsfs}
\usepackage{amssymb}
\usepackage{url}
\usepackage{caption}
\usepackage{enumitem}
\usepackage{mathrsfs}
\usepackage{algorithm}
\usepackage{algpseudocode}
\usepackage{dsfont}
\usepackage{multirow}
\usepackage{booktabs}
\usepackage{array}
\usepackage[pagebackref=false,breaklinks=true,letterpaper=true,colorlinks,urlcolor=magenta,citecolor=blue,linkcolor=blue,bookmarks=false]{hyperref}
\usepackage{colortbl}
\usepackage{nopageno}
\def\cvprPaperID{3141} 

\newcommand{\kk}[1]{\textcolor{black}{#1}}
\newcommand{\figdir}{figures}
\newcommand{\ryn}[1]{\textcolor{black}{#1}}

\ifcvprfinal\fi
\cvprfinalcopy
\begin{document}

\title{Image Correction via Deep Reciprocating HDR Transformation}

\author{Xin Yang$^{2,1\star}$, Ke Xu$^{1,2\star}$, Yibing Song$^{3\dagger}$, Qiang Zhang$^{2}$, Xiaopeng Wei$^{2}$, Rynson~W.H.~Lau$^{1}$ \\
$^{1}$City University of Hong Kong~~$^{2}$Dalian University of Technology~~$^{3}$Tencent AI Lab \\
\small{\url{https://ybsong00.github.io/cvpr18_imgcorrect/index}}
}

\maketitle

\renewcommand{\tabcolsep}{.1pt}
\renewcommand{\thefootnote}{}
\def\swfour{0.245\linewidth}
\begin{figure}[t]
\vspace{-2.8in}
\begin{minipage}{\textwidth}
\begin{center}
\begin{tabular}{cccc}
			\includegraphics[width=\swfour,height=0.79in]{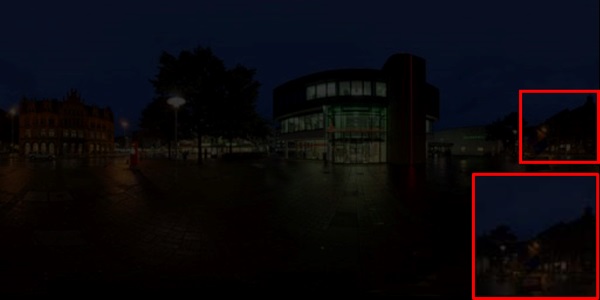}&
			\includegraphics[width=\swfour,height=0.79in]{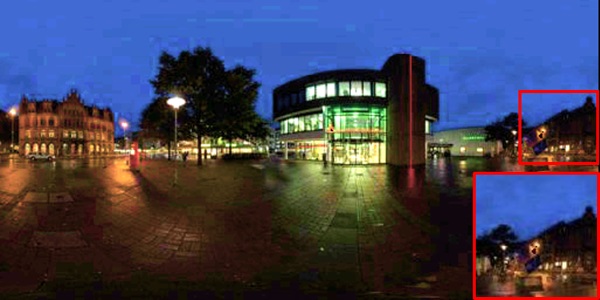}&
			\includegraphics[width=\swfour,height=0.79in]{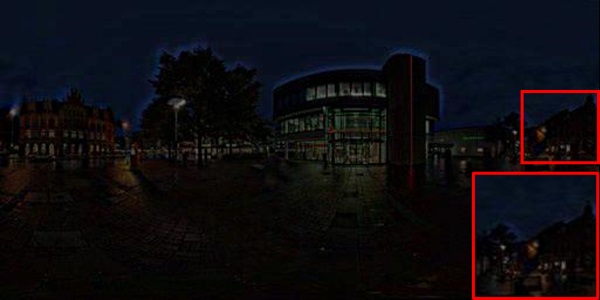}&
			\includegraphics[width=\swfour,height=0.79in]{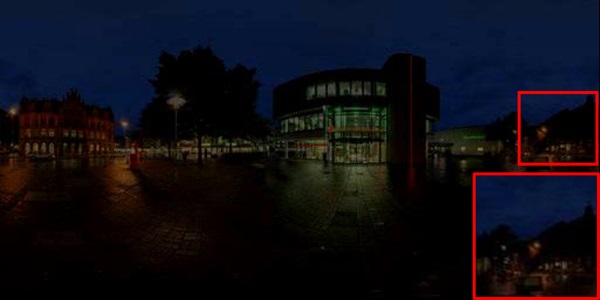}\\
            (a) Input &(b) CAPE~\cite{Kaufman-cgf12-CAPE} &(c) DJF~\cite{Li-ECCV16-DJF} &(d) L0S~\cite{Xu-tog11-L0S} \\
			\includegraphics[width=\swfour,height=0.79in]{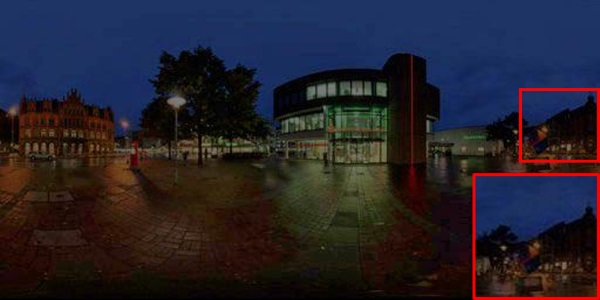}&
			\includegraphics[width=\swfour,height=0.79in]{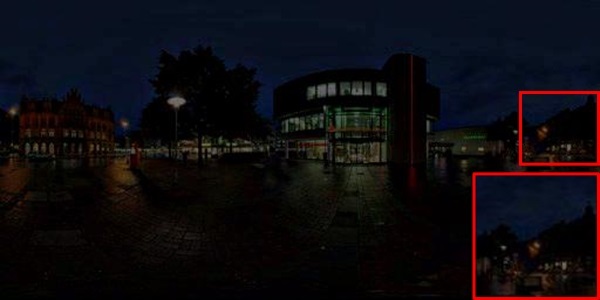}&
			\includegraphics[width=\swfour,height=0.79in]{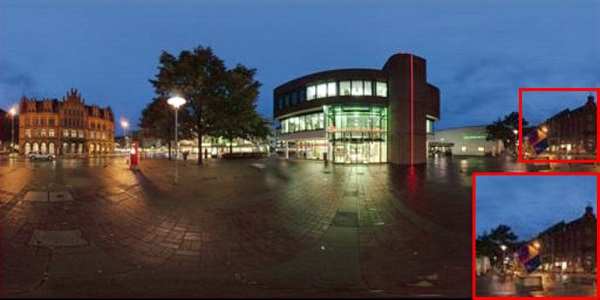}&
			\includegraphics[width=\swfour,height=0.79in]{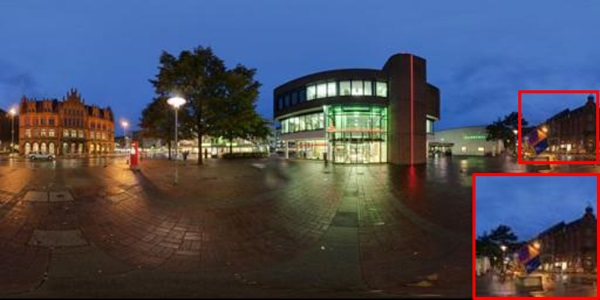}\\
            (e) WVM~\cite{Fu-CVPR16-WVM} &(f) SMF~\cite{Yang-CVPR16-SF} &(g) DRHT &(h) Ground Truth \\

\end{tabular}
\end{center}
\vspace{-6mm}
\caption{Image correction results on an underexposed input. Existing LDR methods have the limitation in recovering the missing details, as shown in (b)-(f). In comparison, we recover the missing LDR details in the HDR domain and preserve them through tone mapping, producing a more favorable result as shown in (g).}
\label{fig:teaser}
\end{minipage}
\end{figure}

\begin{abstract}
Image correction aims to adjust an input image into a visually pleasing one. Existing approaches are proposed mainly from the perspective of image pixel manipulation. They are not effective to recover the details in the under/over exposed regions. In this paper, we revisit the image formation procedure and notice that the missing details in these regions exist in the corresponding high dynamic range (HDR) data. These details are well perceived by the human eyes but diminished in the low dynamic range (LDR) domain because of the tone mapping process. Therefore, we formulate the image correction task as an HDR transformation process and propose a novel approach called \emph{Deep Reciprocating HDR Transformation (DRHT)}. Given an input LDR image, we first reconstruct the missing details in the HDR domain. We then perform tone mapping on the predicted HDR data to generate the output LDR image with the recovered details. To this end, we propose a united framework consisting of two CNNs for HDR reconstruction and tone mapping. They are integrated end-to-end for joint training and prediction. Experiments on the standard benchmarks demonstrate that the proposed method performs favorably against state-of-the-art image correction methods.
\end{abstract}

\section{Introduction}\label{sec:introduction}
The image correction problem has been studied for decades. It dates back to the production of Charge-Coupled Devices (CCDs), which convert optical perception to digital signals. Due to the semiconductors used in the CCDs, there is an unknown nonlinearity existed between the scene radiance and the pixel values in the image. This nonlinearity is usually modeled by gamma correction, which has resulted in a series of image correction methods.
These methods tend to focus on image pixel balance via different approaches including histogram equalization \cite{pizer-CVGIP87-CLAHE}, edge preserving filtering \cite{he-pami13-GF,bao-icpr12-edge}, and CNN encoder-decoder \cite{Tsai-CVPR17-Harmonize}. Typically, they function as a preprocessing step for many machine vision tasks, such as optical flow estimation~\cite{Cheng-ICCV17-SegFlow,Ilg-CVPR17-FlowNet2.0}, image decolorization~\cite{song-wacv14-real,song-siga13-decolor}, image deblurring~\cite{ren-iccv17-video,ren-eccv16-single}, face stylization~\cite{song-ijcai17-faceSketch,song-cviu17-style} and visual tracking~\cite{Song-ICCV17-CREST}.\footnote{$^\star$Joint first authors. $^\dagger$Yibing Song is the corresponding author. This work was conducted at City University of Hong Kong, led by Rynson Lau.}

Despite the demonstrated success, existing methods have the limitation in correcting images with under/over exposure. An example is shown in Figure~\ref{fig:teaser}, where the state-of-the-art image correction methods fail to recover the missing details in the underexposed regions. This is because the pixel values around these regions are close to 0, and the details are diminished within them. Although different image pixel operators have been proposed for image correction, the results are still unsatisfactory, due to the ill-posed nature of the problem. Thus, a question is raised if it is possible to effectively recover the missing details during the image correction process.

To answer the aforementioned question, we trace back to the image formation procedure. Today's cameras still require the photographer to carefully choose the exposure duration ($\Delta t$) and rely on the camera response functions (CRFs) to convert a natural scene ($S$) into an LDR image ($I$), which can be written as~\cite{Debevec-sigg08-crf}:
\begin{equation}
I = f_{CRF}(S \times \Delta t),
\label{eq0:imForm}
\end{equation}
However, when an inappropriate exposure duration is chosen, the existing CRFs can neither correct the raw data in the CCDs nor the output LDR images. This causes the under/over exposure in the LDR images. Based on this observation, we propose an end-to-end framework, called \emph{Deep Reciprocating HDR Transformation (DRHT)}, for image correction. It contains two CNN networks. The first CNN network reconstructs the missing details in the HDR domain and the second CNN network transfers the details back to the LDR domain. Through the reciprocating HDR transformation process, LDR images are corrected in the intermediate HDR domain.

Overall, the contribution in this work can be summarized as follows. We interpret image correction as the Deep Reciprocating HDR Transformation (DRHT) process. An end-to-end DRHT model is therefore proposed to address the image correction problem. To demonstrate the effectiveness of the proposed network, we have conducted extensive evaluations on the proposed network with the state-of-the-art methods, using the standard benchmarks.

\section{Related Work}\label{sec:relatedwork}

In this section, we discuss \ryn{relevant works to our problem}, including \kk{image restoration and filtering, image manipulation, and image enhancement techniques}.

{\flushleft \bf Image Restoration and Filtering.} A variety of state-of-the-art image correction methods have been proposed. Image restoration methods improve the image quality mainly by reducing the \ryn{noise} via different deep network designs~\cite{Kim-CVPR17-DeepAM,Tai-ICCV17-MemNet,Zhang-CVPR17-Deepdenoiser}, low-rank sparse representation learning~\cite{Li-ICCV15-CLSR} or soft-rounding regularization~\cite{Mei-ICCV15-SoftRounding}. \kk{Noise reduction can \ryn{help improve the image quality,} but cannot recover the missing details.} \ryn{Edge-aware} image filtering techniques are also broadly studied for smoothing the images while maintaining \ryn{high} contrasted structures~\cite{Sai-TOG15-L1S,Li-ECCV16-DJF,Shen-ICCV15-MSF}, smoothing repeated textures~\cite{Liu-ECCV16-RNNF,Xu-tog12-structExtract,Yang-CVPR16-SF} or removing high contrast details~\cite{Ma-ICCV13-CTWMF,Zhang-ECCV14-RF,Zhang-CVPR14-fastWMF}. Further operations can be done to enhance the images by strengthening the details filtered out by these methods and then adding them back. Although these filtering methods are sensitive to the local structures, overexposed regions are usually smoothed in the \ryn{output} images and therefore details can hardly be recovered.

{\flushleft \bf Image Manipulation.} Image correction has also been \ryn{done} via pixel manipulation for different \ryn{purposes}, such as color enhancement~\cite{Yan-CVPR14-Learn2Rank} and mimicking different themes/styles~\cite{Wang-tog10-ThemeEnhance,Wang-tog11-StyleEnhance}. Son~\etal~\cite{Son-cgf14-artStyleEnhance} \ryn{propose} a tone transfer model to perform region-dependent tone shifting and scaling for artistic style enhancement. Yan~\etal~\cite{Yan-tog15-deepPhotoAdjust} exploit the image contents and semantics to learn tone adjustments made by photographers via their proposed deep network. However, these works \ryn{mainly focus} on manipulating the LDR images to \ryn{adapt to various user preferences}.

{\flushleft \bf Image Enhancement.} Histogram equalization is the most widely used method for image enhancement by balancing the histogram of the image. Global and local contrast adjustments are also studied in~\cite{Hwang-ECCV12-contextEnhance,Rivera-tip12-ChannelDivEnhance} for enhancing the contrast and brightness. Kaufman~\etal~\cite{Kaufman-cgf12-CAPE} propose a framework to apply carefully designed operators to strengthen the detected regions (\eg, faces and skies), in addition to the global contrast and saturation manipulation. Fu~\etal~\cite{Fu-CVPR16-WVM} propose a weighted variational method to jointly estimate the reflectance and illumination for color correction. Guo~\etal~\cite{Guo-tip17LIME} propose to first reconstruct and refine the illumination map from the maximum values in the RGB channels and then enhance the illumination map. Recently, Shen~\etal~\cite{Shen-arXiv17-MSRnet} propose a deep network to directly learn the mapping relations of low-light and ground truth images. This method can successfully recover rich details buried in low light conditions, but \ryn{it} tends to increase the global illumination and generate surrealistic images.

\ryn{All these} methods, however, cannot completely recover the missing details in the bright and dark regions. This is mainly because both their inputs and their enhancing operations are restricted to work in the LDR domain, which does not offer sufficient information to recover all the details while maintaining the global illumination.

\def\swone{0.9\linewidth}
\begin{figure*}[t]
	\begin{center}
    \begin{tabular}{c}
            \vspace{-2mm}\includegraphics[width=0.8\linewidth]{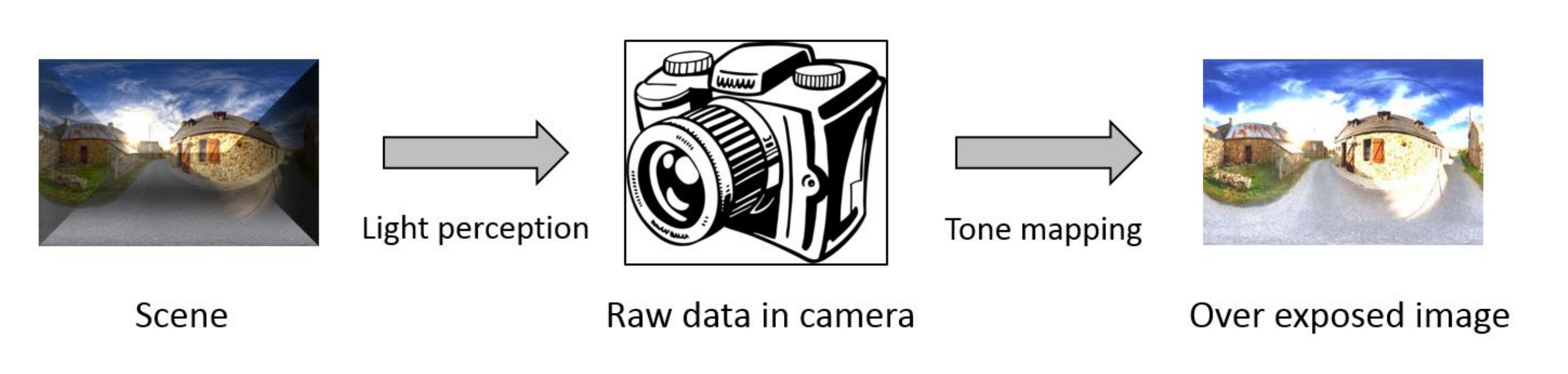}\\
            (a) Image formulation process \\
            \includegraphics[width=\swone]{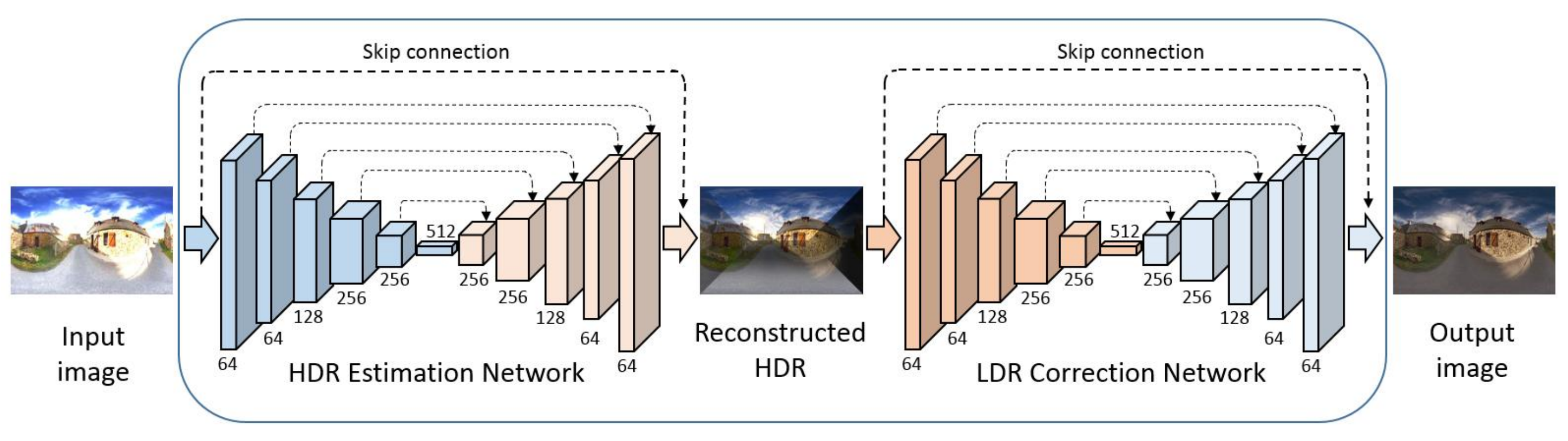}\\
            (b) Deep Reciprocating HDR Transformation (DRHT) pipeline\\
    \end{tabular}
    \end{center}
    \caption{An overview of image formulation process and the proposed DRHT pipeline. Given an input under/over exposed LDR image, we first reconstruct the missing details in the HDR domain and map them back to the output LDR domain. }
	\label{fig:overview}
\end{figure*}

\section{Deep Reciprocating HDR Transformation}
An overview of the proposed method is shown in Figure~\ref{fig:overview}(b). We first illustrate our reformulation of image correction. We then show our HDR estimation network to predict HDR data given LDR input. Finally, we show that the HDR data is tone mapped into the output LDR using a LDR correction network. The details are presented as follows:

\subsection{Image Correction Reformulation}
Although human can well perceive the HDR data, it requires empirically configuring the camera during the imaging process. An overview of scene capturing and producing LDR is shown in Figure~\ref{fig:overview}(a). However, when under extreme lighting conditions (e.g., the camera is facing the sun), details in the natural scenes are lost during the tone mapping process. They cannot be recovered by existing image correction methods in the LDR domain.

In order to recover the degraded regions caused by under/over exposures, we trace back to the image formation procedure and formulate the correction as the Deep Reciprocating HDR Transformation process:
$\hat{S} = f_1(I; \theta_{1})$ and $\hat{I}^{ldr} = f_2(\hat{S}; \theta_{2})$,
where $\hat{S}$ and $\hat{I}^{ldr}$ represent the reconstructed HDR data and the corrected LDR image, respectively. $\theta_{1}$ and $\theta_{2}$ are the CNN parameters. Specifically, we propose the HDR estimation network ($f_1$) to first recover the details in the HDR domain and then the LDR correction network ($f_2$) to transfer the recovered HDR details back to the LDR domain. Images are corrected via this end-to-end DRHT process.

\subsection{HDR Estimation Network}
We propose a HDR estimation network to recover the missing details in the HDR domain, as explained below:

{\flushleft \bf Network Architecture.}
Our network is based on a fully convolutional encoder-decoder network. Given an input LDR image, we encode it into a low dimensional latent representation, which is then decoded to reconstruct the HDR data. Meanwhile, we add skip connections from each encoder layer to its corresponding decoder layer. They enrich the local details during decoding in a coarse-to-fine manner. To facilitate the training process, we also add a skip connection directly from the input LDR to the output HDR. Instead of learning to predict the whole HDR data, the HDR estimation network only needs to predict the difference between the input and output, which shares some similarity to residual learning~\cite{He-cvpr16-resnet}. We train this network from scratch and use batch normalization~\cite{Toffe-ICML15-BatchNorm} and ELU~\cite{Clevert-arXiv15-elu} activation for all the convolutional layers.

{\flushleft \bf Loss Function.}
Given an input image $I$, the output of this network $\hat{S} = f_{1}(I;\theta_1)$, and the ground truth HDR image $Y$, we use the Mean Square Error (MSE) as the objective function:
\begin{equation}
Loss_{hdr} = \frac{1}{2N}\sum_{i=1}^{N}\left \| \hat{S_i} - \alpha(Y_i)^{\gamma} \right \|_2^2,
\label{eq1:hdrloss}
\end{equation}
where $i$ is the pixel index and $N$ refers to the total number of pixels. $\alpha$ and $\gamma$ are two constants in the nonlinear function to convert the ground truth HDR data into LDR, which is empirically found to facilitate the network convergence. We pretrain this network in advance before integrating it with the remaining modules.

\subsection{LDR Correction Network}
We propose a LDR correction network, which shares the same architecture as that of the HDR estimation network. It aims to preserve the recovered details in the LDR domain, as explained below:

{\flushleft \bf Loss Function.}
The output of the HDR estimation network $\hat{S}$ is in LDR as shown in Eq. \ref{eq1:hdrloss}. We first map it to the HDR domain via inverse gamma correction. The mapped result is denoted as $\hat{S}_{full}$. We then apply a logarithmic operation to preserve the majority of the details and feed the output to the LDR correction network. Hence, the recovered LDR image $\hat{I}^{ldr}$ through our network becomes:
\begin{equation}
\hat{I}^{ldr} = f_{2}(log(\hat{S}_{full}+\delta);\theta_2),
\label{eq2:trans}
\end{equation}
where $log()$ is used to compress the full HDR domain for convergence while maintaining a relatively large range of intensity, and $\delta$ is a small constant to remove zero values. With the ground truth LDR image $I^{gt}$, the loss function is:
\begin{equation}
Loss_{ldr} = \frac{1}{2N}\sum_{i=1}^{N}(\left \| \hat{I_i}^{ldr} - I_{i}^{gt} \right \|_2^2 + \epsilon \left \| \hat{S_i} - \alpha(Y_i)^{\gamma} \right \|_2^2),
\label{eq3:ldrloss}
\end{equation}
where $\epsilon$ is a balancing parameter to control the influence of the HDR reconstruction accuracy.

{\flushleft \bf Hierarchical Supervision.}
\kk{We train this LDR correction network together with the aforementioned HDR estimation network. We adopt this end-to-end training strategy in order to adapt our whole model to the domain reciprocating transformation.} To facilitate the training process, we adopt the hierarchical supervision training strategies similar to~\cite{He-ICCV17-salDetect}. Specifically, we start to train the encoder part and the shallowest deconv layer of the LDR correction network by freezing the learning rates of all other higher deconv layers. During training, higher deconv layers are gradually added for fine tuning while the learning rates of the encoder and shallower deconv layers will be decreased. In this way, this network can learn to transfer the HDR details to LDR domain in a coarse-to-fine manner.

\subsection{Implementation Details}
The proposed DRHT model is implemented under the Tensorflow framework~\cite{Tensorflow} on a PC with an i7 4GHz CPU and an NVIDIA GTX 1080 GPU. The network parameters are initialized using the truncated normal initializer. We use $9\times9$ and $5\times5$ kernel sizes to generate 64-dimensional feature maps for the first two conv layers and their counterpart deconv layers for both networks, and the remaining kernel size is set to $3\times3$.  For loss minimization, we adopt the ADAM optimizer~\cite{Kingma-arXiv14-Adam} with an initial learning rate of 1e-2 for 300 epochs, and then use learning rate of 5e-5 with momentum $\beta_1 = 0.9$ and $\beta_2 = 0.998$ for another 100 epochs. $\alpha$ and $\gamma$ in Eq.~\ref{eq1:hdrloss}, and $\delta$ in Eq.~\ref{eq2:trans} are set to $0.03$, $0.45$ and $1/255$, respectively. We also clip the gradients to avoid the gradient explosion problem. The general training takes about ten days and the test time is about 0.05s for a 256$\times$512 image.

\section{Experiments}

In this section, we first present the experiment setups and internal analysis on the effectiveness of the HDR estimation network. We then compare our DRHT model with the state-of-the-art image correction methods on two datasets.

\def\swfour{0.24\linewidth}
\begin{figure}[t]
	\begin{center}
\begin{small}
    \begin{tabular}{cccc}
			\includegraphics[width=\swfour]{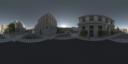}&
			\includegraphics[width=\swfour]{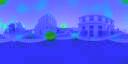}&
            \includegraphics[width=\swfour]{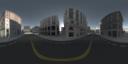}&
            \includegraphics[width=\swfour]{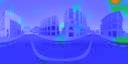}\\
            (a) Input & DRHT (64.75) &(b) Input & DRHT (65.61) \\
			\includegraphics[width=\swfour]{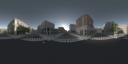}&
			\includegraphics[width=\swfour]{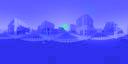}&
            \includegraphics[width=\swfour]{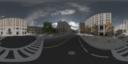}&
            \includegraphics[width=\swfour]{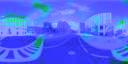}\\
            (c) Input & DRHT (61.80) &(d) Input & DRHT (69.28) \\
			\includegraphics[width=\swfour]{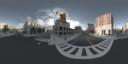}&
			\includegraphics[width=\swfour]{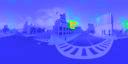}&
            \includegraphics[width=\swfour]{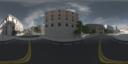}&
            \includegraphics[width=\swfour]{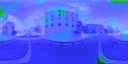}\\
            (e) Input & DRHT (62.69) &(f) Input & DRHT (69.04) \\
			\includegraphics[width=\swfour]{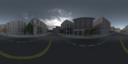}&
			\includegraphics[width=\swfour]{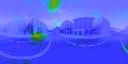}&
            \includegraphics[width=\swfour]{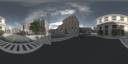}&
            \includegraphics[width=\swfour]{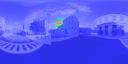}\\
            (g) Input & DRHT (69.57) &(h) Input & DRHT (62.17) \\
            \includegraphics[width=\swfour]{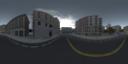}&
			\includegraphics[width=\swfour]{\figdir/hdrAnalys/P_11_Q_61_8028.jpg}&
            \includegraphics[width=\swfour]{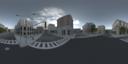}&
            \includegraphics[width=\swfour]{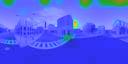}\\
            \vspace{2mm}(i) Input & DRHT (61.80) &(j) Input & DRHT (65.18) \\
    \end{tabular}
    \includegraphics[width=0.7\linewidth]{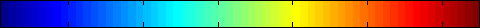}\\
    low difference~~~~~~~~~~~~~~~~~~~~~~~~~~~~~~~~~~~~~high difference\\
    \end{small}
    \end{center}
    \vspace{-5mm}
    \caption{Internal Analysis. We compare the reconstructed HDR images with the ground truth HDR images using the HDR-VDP-2 metric. The average Q score and SSIM index on this test set are $61.51$ and $0.9324$, respectively.}
	\label{fig:hdrAnalys}
\end{figure}

\def\swfour{0.245\linewidth}
\begin{figure*}[t]
	\begin{center}
    \begin{tabular}{cccc}
			\includegraphics[width=\swfour,height=0.70in]{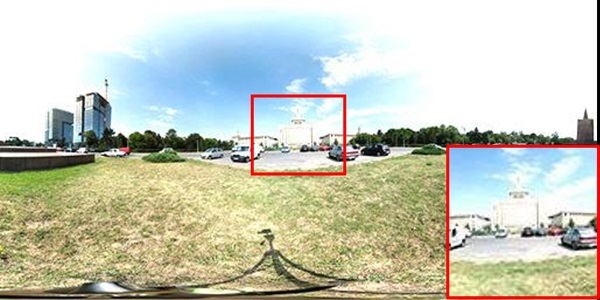}&
			\includegraphics[width=\swfour,height=0.70in]{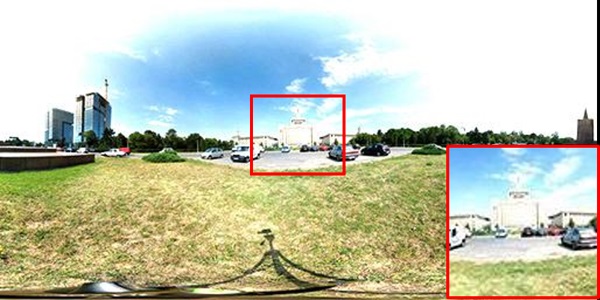}&
			\includegraphics[width=\swfour,height=0.70in]{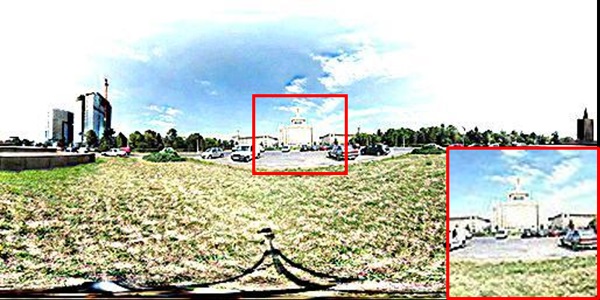}&
			\includegraphics[width=\swfour,height=0.70in]{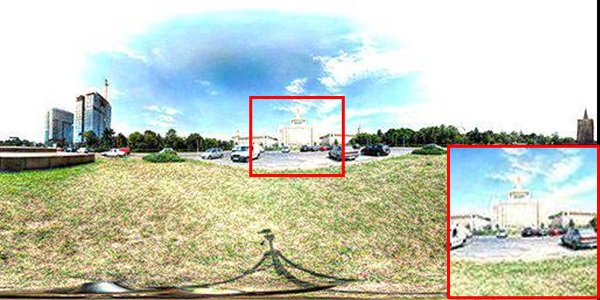}\\
            (a) Input &(b) CAPE~\cite{Kaufman-cgf12-CAPE} &(c) DJF~\cite{Li-ECCV16-DJF} &(d) L0S~\cite{Xu-tog11-L0S} \\
			\includegraphics[width=\swfour,height=0.70in]{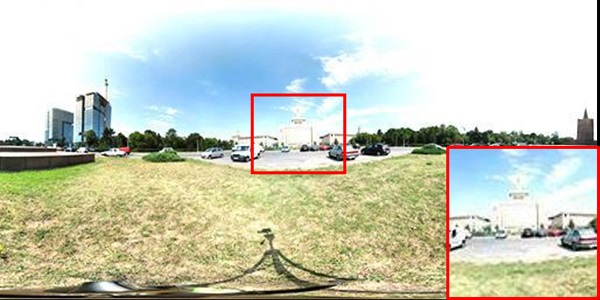}&
			\includegraphics[width=\swfour,height=0.70in]{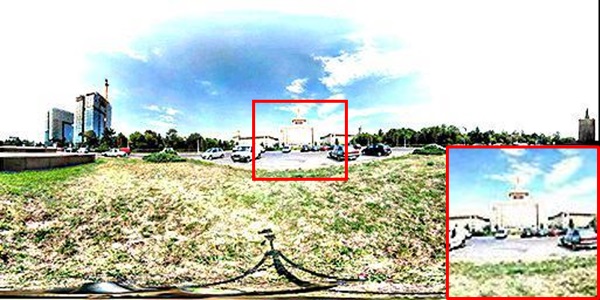}&
			\includegraphics[width=\swfour,height=0.70in]{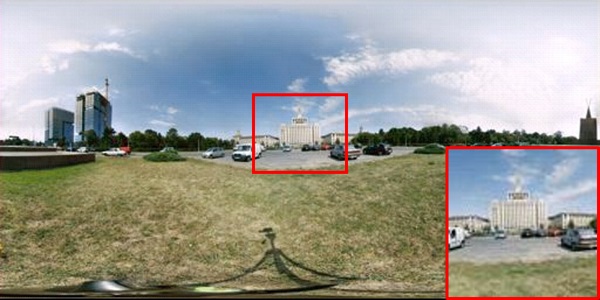}&
			\includegraphics[width=\swfour,height=0.70in]{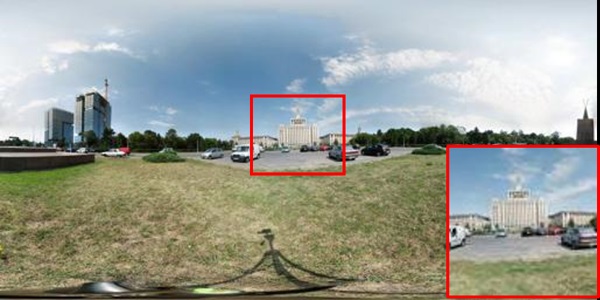}\\
            (e) WVM~\cite{Fu-CVPR16-WVM}  &(f) SMF~\cite{Yang-CVPR16-SF} &(g) DRHT &(h) Ground Truth  \\
			\includegraphics[width=\swfour,height=0.70in]{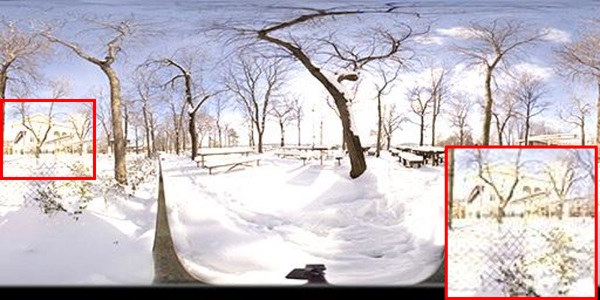}&
			\includegraphics[width=\swfour,height=0.70in]{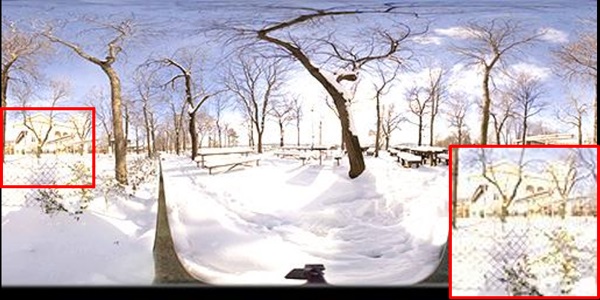}&
			\includegraphics[width=\swfour,height=0.70in]{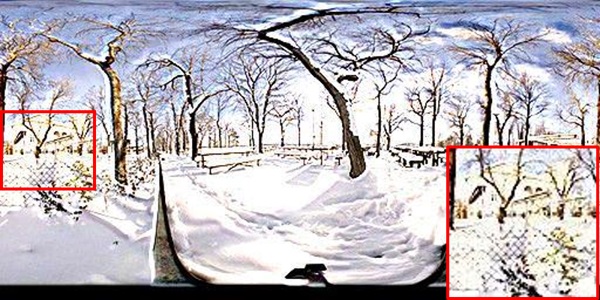}&
			\includegraphics[width=\swfour,height=0.70in]{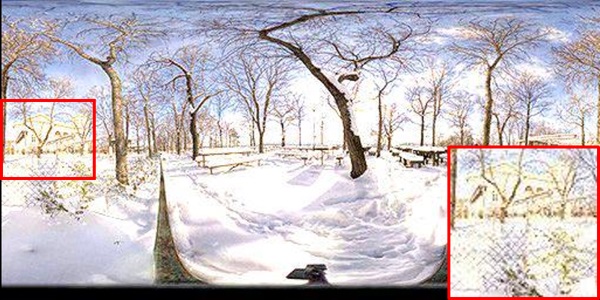}\\
            (i) Input &(j) CAPE~\cite{Kaufman-cgf12-CAPE} &(k) DJF~\cite{Li-ECCV16-DJF} &(l) L0S~\cite{Xu-tog11-L0S} \\
			\includegraphics[width=\swfour,height=0.70in]{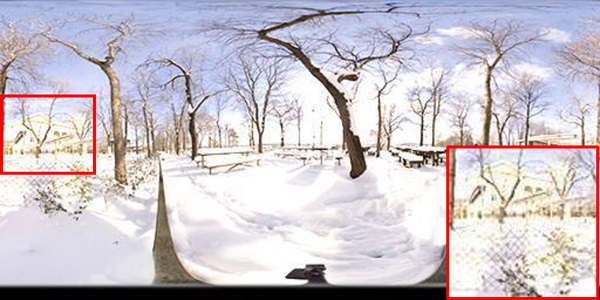}&
			\includegraphics[width=\swfour,height=0.70in]{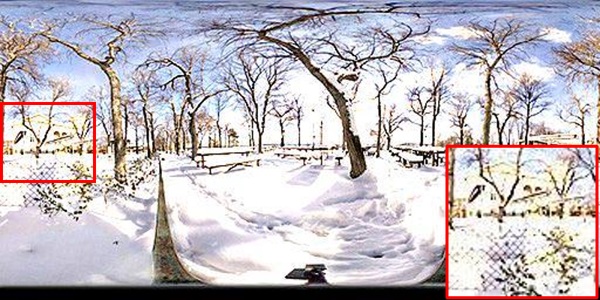}&
			\includegraphics[width=\swfour,height=0.70in]{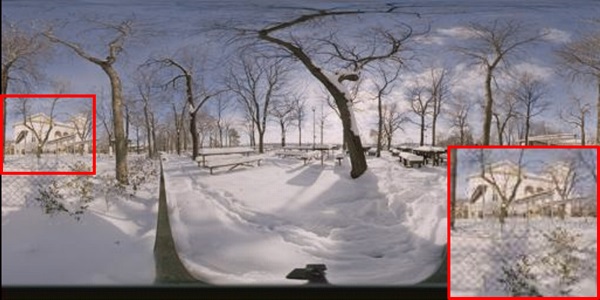}&
			\includegraphics[width=\swfour,height=0.70in]{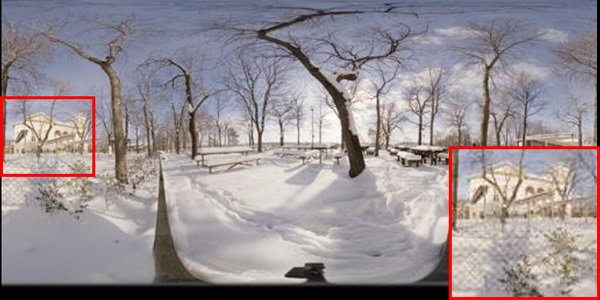}\\
            (m) WVM~\cite{Fu-CVPR16-WVM} &(n) SMF~\cite{Yang-CVPR16-SF} &(o) DRHT &(p) Ground Truth \\
            \includegraphics[width=\swfour,height=0.70in]{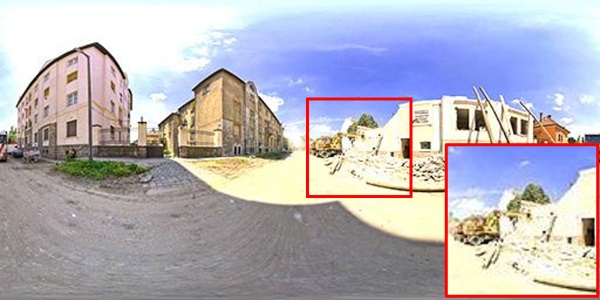}&
            \includegraphics[width=\swfour,height=0.70in]{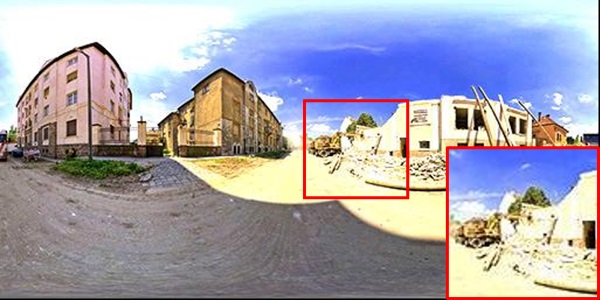}&
            \includegraphics[width=\swfour,height=0.70in]{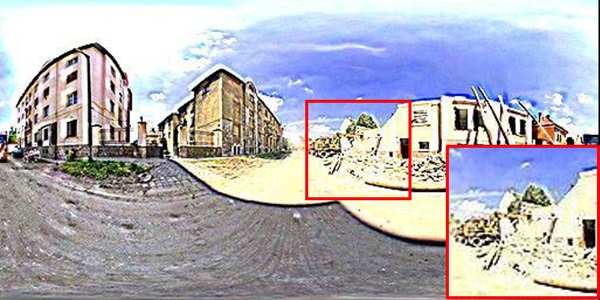}&
            \includegraphics[width=\swfour,height=0.70in]{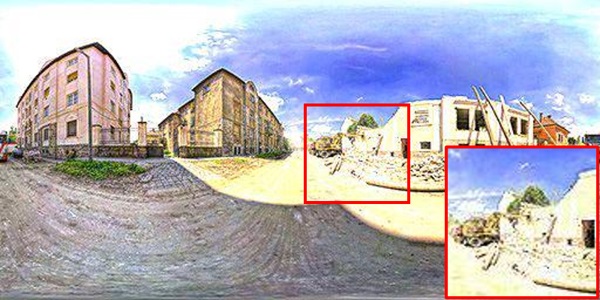}\\
            (q) Input &(r) CAPE~\cite{Kaufman-cgf12-CAPE} &(s) DJF~\cite{Li-ECCV16-DJF} &(t) L0S~\cite{Xu-tog11-L0S} \\
            \includegraphics[width=\swfour,height=0.70in]{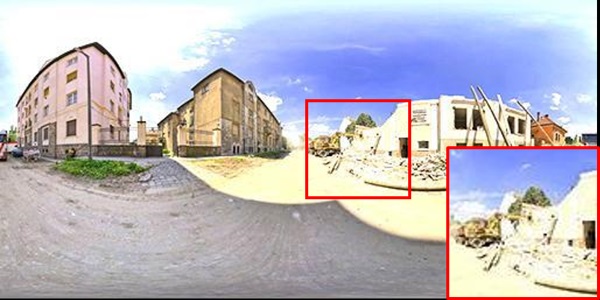}&
            \includegraphics[width=\swfour,height=0.70in]{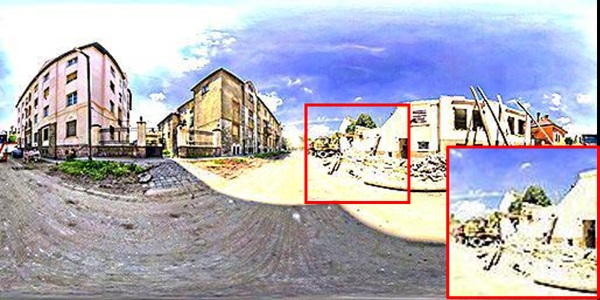}&
            \includegraphics[width=\swfour,height=0.70in]{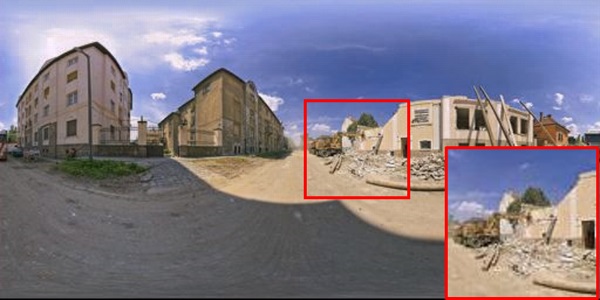}&
            \includegraphics[width=\swfour,height=0.70in]{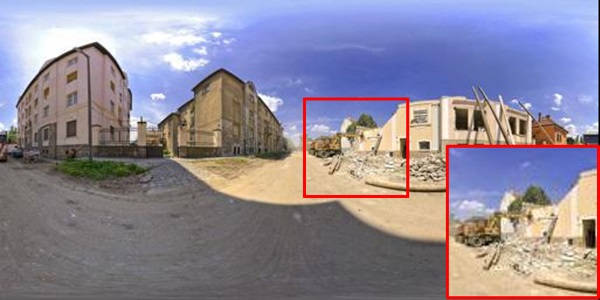}\\
            (u) WVM~\cite{Fu-CVPR16-WVM} &(v) SMF~\cite{Yang-CVPR16-SF} &(w) DRHT &(x) Ground Truth \\
    \end{tabular}
    \end{center}
    \vspace{-3mm}
    \caption{Visual comparison on overexposed images in the bright scenes. The proposed DRHT method can effectively recover the missing details buried in the overexposed regions compared with state-of-the-art approaches}
	\label{fig:overCompare}
\end{figure*}

\subsection{Experiments Setups}

{\flushleft \bf Datasets.}
We conduct experiments on the city scene panorama dataset~\cite{Zhang-ICCV17-ldr2hdr} and the Sun360 outdoor panorama dataset~\cite{Xiao-CVPR12-SUN360}. Specifically, since the low-resolution (64$\times$128 pixels) city scene panorama dataset~\cite{Zhang-ICCV17-ldr2hdr} contains LDR and ground truth HDR image pairs, we use the black-box Adobe Photoshop software to empirically generate ground truth LDR images with human supervision. Therefore, we use $39,198$ image pairs (i.e., the input LDR and the ground truth HDR) to train the first network and use $39,198$ triplets (i.e., the input LDR, the ground truth HDR and the ground truth LDR) to train the whole network. We use $1,672$ images from their testing set for evaluation. To adapt our models to the real images with high resolution, we use the Physically Based Rendering Technology (PBRT)~\cite{PBRT} to generate $119$ ground truth HDR scenes as well as the input and ground truth LDR images, which are then divided into $42,198$ patches for training. We also use $6,400$ images from the Sun360 outdoor panorama dataset~\cite{Xiao-CVPR12-SUN360} for end-to-end finetuning (i.e., $\epsilon$ in Eq.~\ref{eq3:ldrloss} is fixed as $0$), as they do not have ground truth HDR images, and use $1,200$ images for evaluation. The input images are corrupted from the originals by adjusting the exposure (selected from the interval [-6, 3], in order not to learn the mapping between one specific exposure degree and the ground truth) and contrasts to over/under expose the visible details. We resize the images to 256$\times$512 pixels in this dataset.

\def\swfour{0.245\linewidth}
\begin{figure*}
	\begin{center}
    \begin{tabular}{cccc}
			\includegraphics[width=\swfour,height=0.70in]{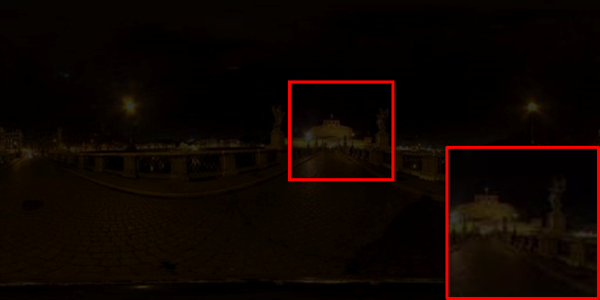}&
			\includegraphics[width=\swfour,height=0.70in]{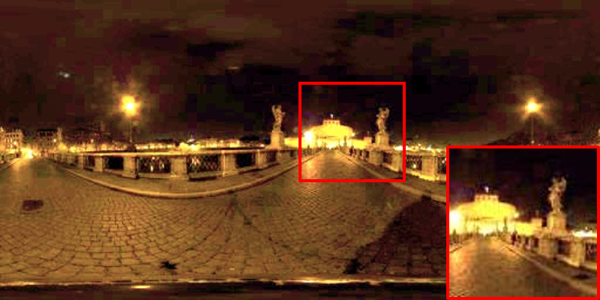}&
			\includegraphics[width=\swfour,height=0.70in]{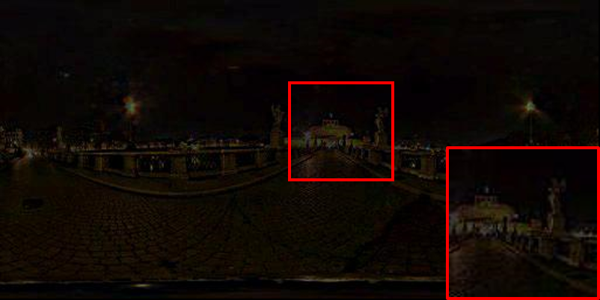}&
			\includegraphics[width=\swfour,height=0.70in]{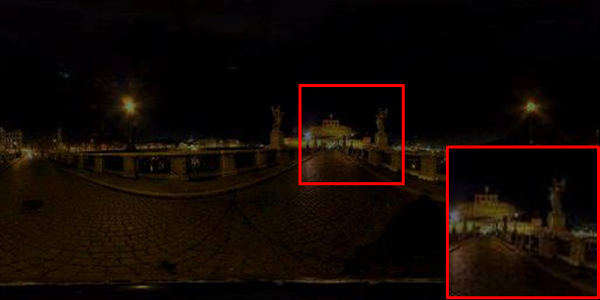}\\
            (a) Input &(b) CAPE~\cite{Kaufman-cgf12-CAPE} &(c) DJF~\cite{Li-ECCV16-DJF} &(d) L0S~\cite{Xu-tog11-L0S} \\
			\includegraphics[width=\swfour,height=0.70in]{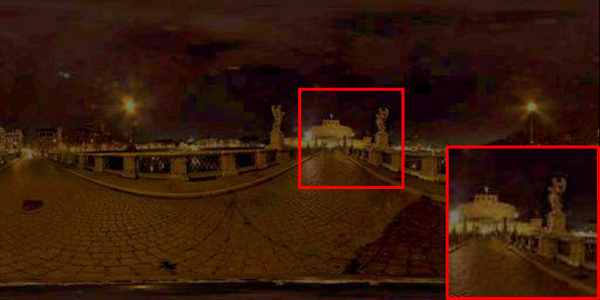}&
			\includegraphics[width=\swfour,height=0.70in]{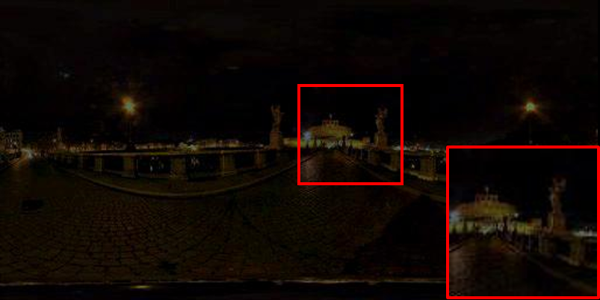}&
			\includegraphics[width=\swfour,height=0.70in]{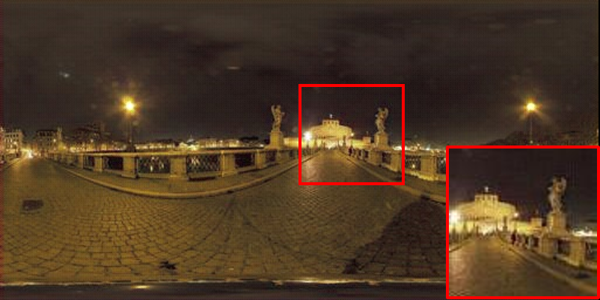}&
			\includegraphics[width=\swfour,height=0.70in]{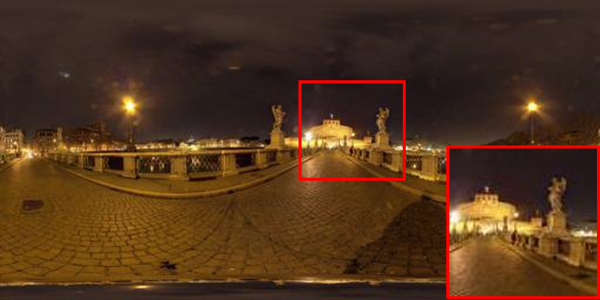}\\
            (e) WVM~\cite{Fu-CVPR16-WVM} &(f) SMF~\cite{Yang-CVPR16-SF} &(g) DRHT &(h) Ground Truth \\
            \includegraphics[width=\swfour,height=0.70in]{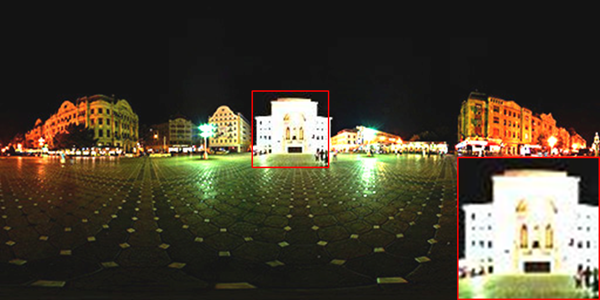}&
            \includegraphics[width=\swfour,height=0.70in]{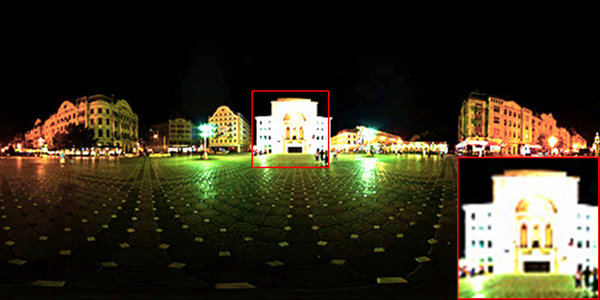}&
            \includegraphics[width=\swfour,height=0.70in]{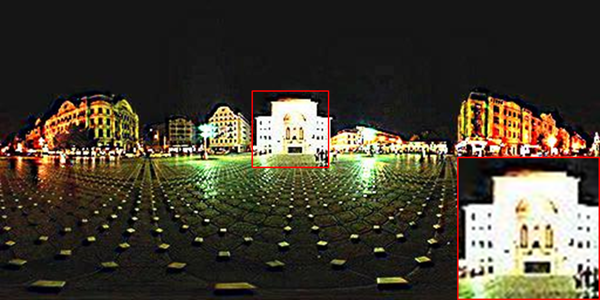}&
            \includegraphics[width=\swfour,height=0.70in]{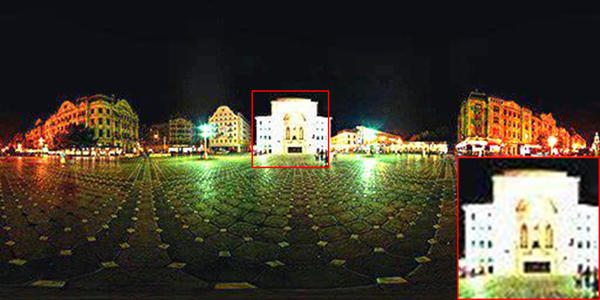}\\
            (i) Input &(j) CAPE~\cite{Kaufman-cgf12-CAPE} &(k) DJF~\cite{Li-ECCV16-DJF} &(l) L0S~\cite{Xu-tog11-L0S} \\
            \includegraphics[width=\swfour,height=0.70in]{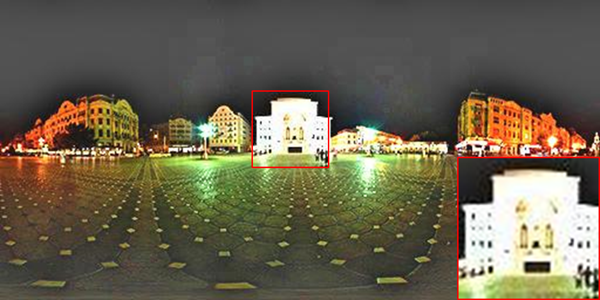}&
            \includegraphics[width=\swfour,height=0.70in]{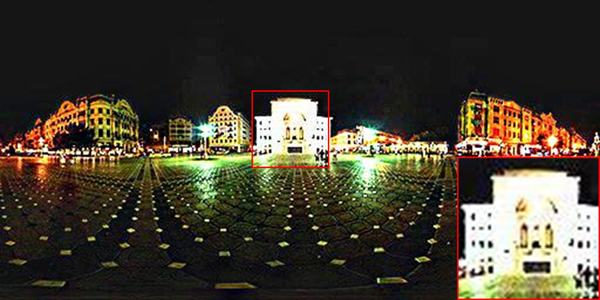}&
            \includegraphics[width=\swfour,height=0.70in]{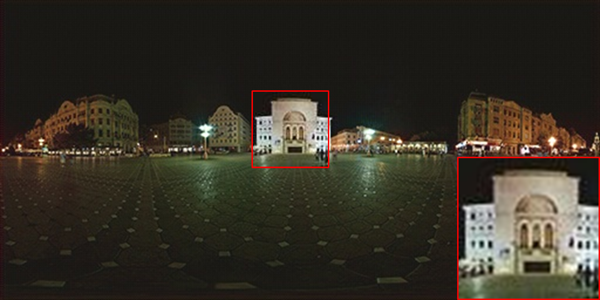}&
            \includegraphics[width=\swfour,height=0.70in]{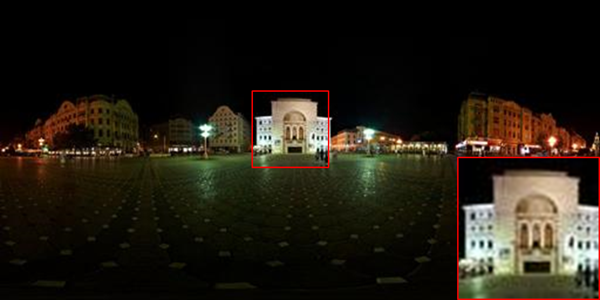}\\
            (m) WVM~\cite{Fu-CVPR16-WVM} &(n) SMF~\cite{Yang-CVPR16-SF} &(o) DRHT &(p) Ground Truth \\
            \includegraphics[width=\swfour,height=0.70in]{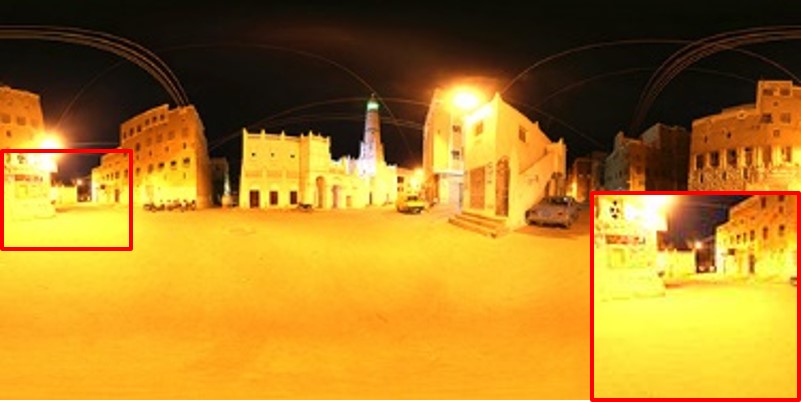}&
            \includegraphics[width=\swfour,height=0.70in]{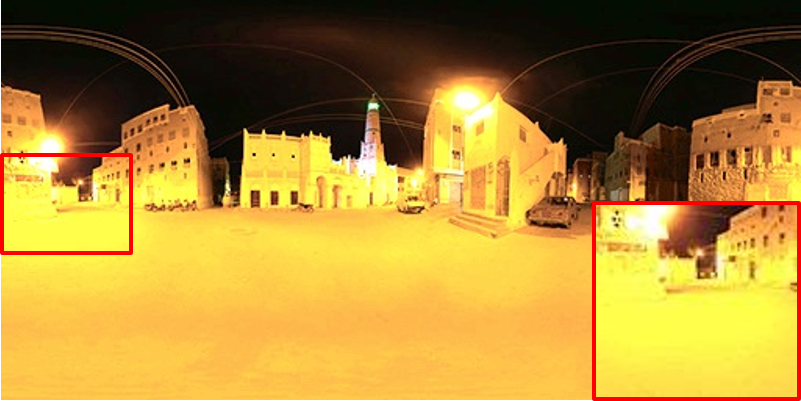}&
            \includegraphics[width=\swfour,height=0.70in]{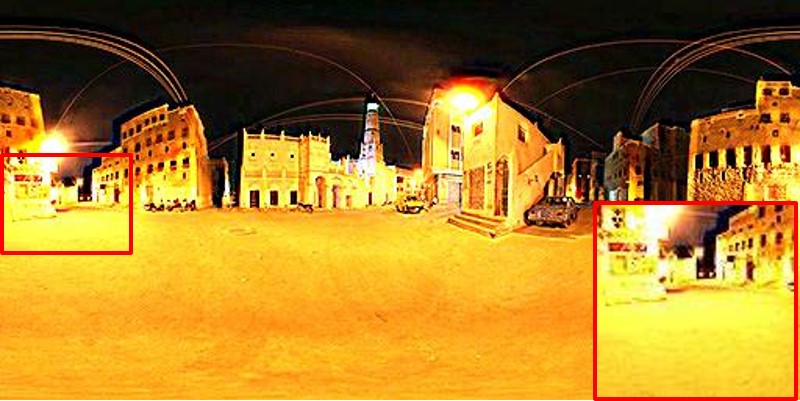}&
            \includegraphics[width=\swfour,height=0.70in]{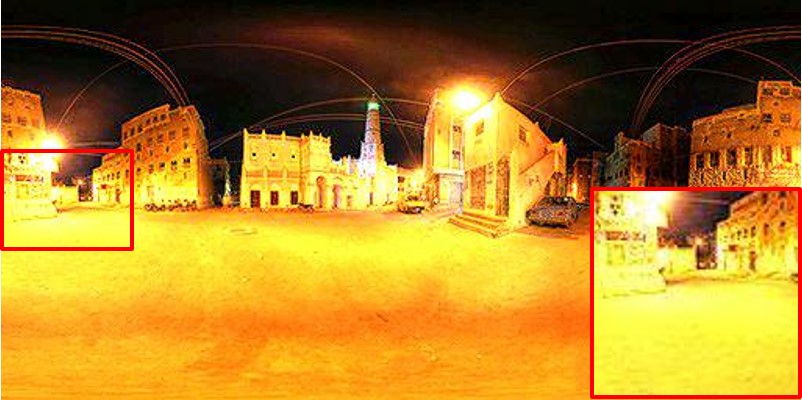}\\
            (q) Input &(r) CAPE~\cite{Kaufman-cgf12-CAPE} &(s) DJF~\cite{Li-ECCV16-DJF} &(t) L0S~\cite{Xu-tog11-L0S} \\
            \includegraphics[width=\swfour,height=0.70in]{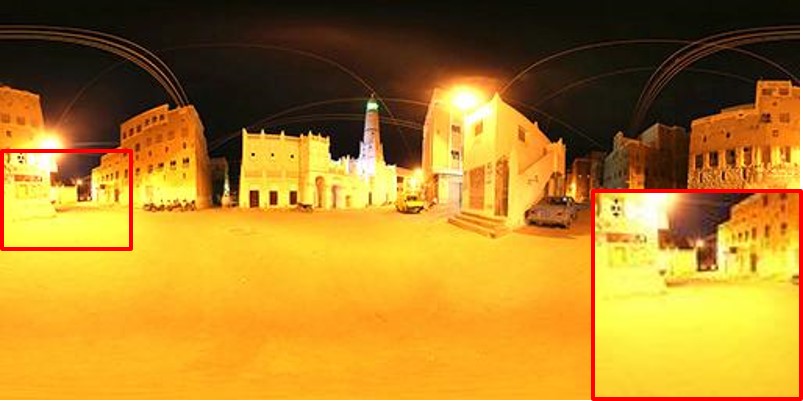}&
            \includegraphics[width=\swfour,height=0.70in]{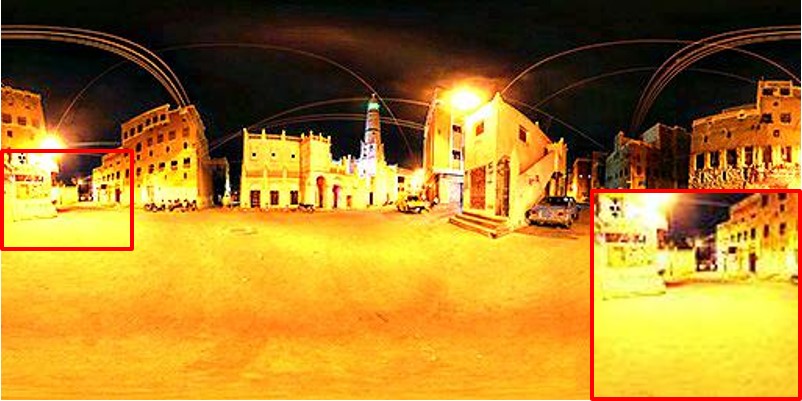}&
            \includegraphics[width=\swfour,height=0.70in]{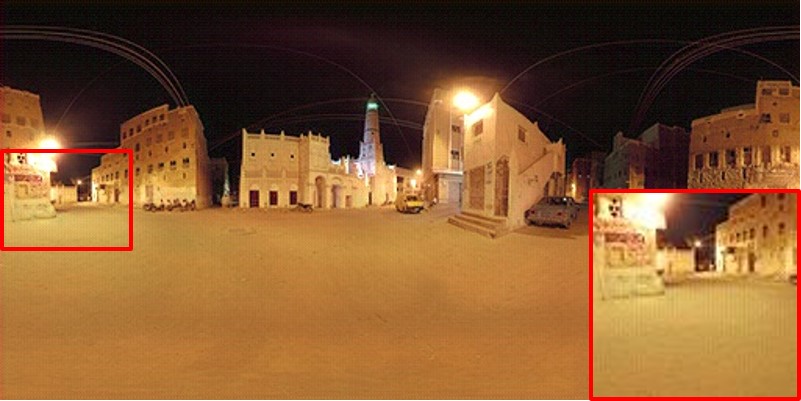}&
            \includegraphics[width=\swfour,height=0.70in]{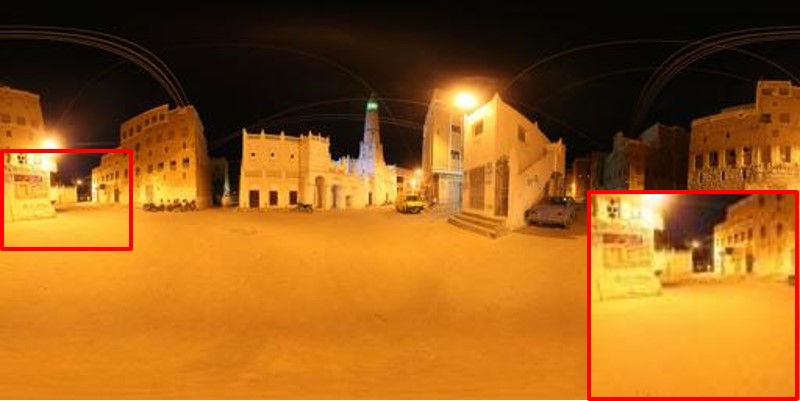}\\
            (u) WVM~\cite{Fu-CVPR16-WVM} &(v) SMF~\cite{Yang-CVPR16-SF} &(w) DRHT &(x) Ground Truth \\
    \end{tabular}
    \end{center}
    \vspace{-3mm}
    \caption{Visual comparison on under/over exposed images in the dark scenes. The proposed DRHT method can effectively recover the missing details in the under/over exposed regions while maintaining the global illumination.}
	\label{fig:underCompare}
\end{figure*}

{\flushleft \bf Evaluation Methods.}
We compare the proposed method to $5$ state-of-the-art image correction methods Cape~\cite{Kaufman-cgf12-CAPE},WVM~\cite{Fu-CVPR16-WVM}, SMF~\cite{Yang-CVPR16-SF}, L0S~\cite{Xu-tog11-L0S} and DJF~\cite{Li-ECCV16-DJF} on the dataset. Among them, Cape~\cite{Kaufman-cgf12-CAPE} enhances the images via a comprehensive pipeline including global contrast/saturation correction, sky/face enhancement, shadow-saliency and texture enhancement. WVM~\cite{Fu-CVPR16-WVM} first decomposes the input image into reflectance and illumination maps, and corrects the input by enhancing the illumination map. Since the enhancement operations are mostly conducted on the detail layer extracted by existing filtering methods, we further compare our results to state-of-the-art image filtering methods. Meanwhile, we compare the proposed method to two deep learning based image correction methods: Hdrcnn~\cite{Eilertsen-tog17-hdrcnn} and DrTMO~\cite{Endo-tog17-DrTMO}.

{\flushleft \bf Evaluation Metrics.}
We evaluate the performance using different metrics. When internal analyzing the HDR estimation network, we use the widely adopted HDR-VDP-2~\cite{Mantiuk-tog11-hdrvdp2} metric it reflects human perception on different images. When comparing with existing methods, we use three commonly adopted image quality metrics: PSNR, SSIM~\cite{SSIM} and FSIM~\cite{FSIM}. In addition, we provide the Q scores from the HDR-VDP-2~\cite{Mantiuk-tog11-hdrvdp2} metric to evaluate the image quality.

\subsection{Internal Analysis}
As the proposed DRHT method first recovers the details via the HDR estimation network, we demonstrate its effectiveness in reconstructing the details in the HDR domain. We evaluate on the city scene dataset using the HDR-VDP-2 metric~\cite{Mantiuk-tog11-hdrvdp2}. It generates the probability map and the Q score for each test image. The probability map indicates the difference between two images to be noticed by an observer on average. Meanwhile, the Q score predicts the quality degradation through a Mean-Opinion-score metric.

We provide some examples in Figure~\ref{fig:hdrAnalys} which are from the city scene test dataset. We overlay the predicted visual difference on the generated result. The difference intensity is shown via a color bar where the low intensity is marked as blue while the high intensity is marked as red. It shows that the proposed HDR estimation network can effectively recover the missing details on the majority of the input image. However, the limitation appears on the region where the part of sun is occluded by the building, as shown in (j). It brings high difference because the illumination contrast is high around the boundary between sun and the building. This difference is difficult to preserve in the HDR domain. The average Q score and SSIM index on this test set are $61.51$ and $0.9324$, respectively. They indicate that the synthesized HDR data through our HDR estimation network is close to the ground truth HDR data.

\begin{table*}[t]
\begin{center}
\begin{tabular}{|p{1.8cm}<{\centering}|p{1.2cm}<{\centering}p{1.2cm}<{\centering}p{1.2cm}<{\centering}p{1.2cm}<{\centering}|p{1.2cm}<{\centering}p{1.2cm}<{\centering}p{1.2cm}<{\centering}p{1.2cm}<{\centering}|}
\hline
\multirow{2}{*}{Methods} &\multicolumn{4}{c|}{City Scene dataset} &\multicolumn{4}{c|}{Sun360 Outdoor dataset} \\
\cline{2-9}
            & PSNR & SSIM &FSIM &Q score  & PSNR & SSIM &FSIM &Q score  \\
\hline
CAPE~\cite{Kaufman-cgf12-CAPE}         &18.99    &0.7435    &0.8856     &59.44         &17.13    &0.7853     &0.8781     &54.87        \\
WVM~\cite{Fu-CVPR16-WVM}               &17.70    &0.8016    &0.8695     &53.17         &11.25    &0.5733     &0.6072     &41.12        \\
L0S~\cite{Xu-tog11-L0S}                &19.03    &0.6644    &0.7328     &84.33         &15.72    &0.7311     &0.7751     &51.73        \\
SMF~\cite{Yang-CVPR16-SF}              &18.61    &0.7724    &0.9035     &81.07         &14.85    &0.6776     &0.7622     &50.77        \\
DJF~\cite{Li-ECCV16-DJF}               &17.54    &0.7395    &0.9512     &84.74         &14.49    &0.6736     &0.7360     &50.03        \\
\rowcolor[gray]{.7} DRHT                                   &28.18    &0.9242    &0.9622     &97.87         &22.60    &0.7629     &0.8691     &56.17        \\
\hline
\end{tabular}
\end{center}
\vspace{-3.0mm}
\caption{Quantitative evaluation on the standard datasets. The proposed DRHT method is compared with existing image correction methods based on several metrics including PSNR, SSIM, FSIM and Q score. It shows that the proposed DRHT method performs favorably against existing image correction methods.}
\label{tab:psnr}
\end{table*}

\begin{table*}
\begin{center}
\begin{tabular}{|p{1.8cm}<{\centering}|p{1.2cm}<{\centering}p{1.2cm}<{\centering}p{1.2cm}<{\centering}p{1.2cm}<{\centering}|p{1.2cm}<{\centering}p{1.2cm}<{\centering}p{1.2cm}<{\centering}p{1.2cm}<{\centering}|}
\hline
\multirow{2}{*}{Methods} &\multicolumn{4}{c|}{City Scene dataset} &\multicolumn{4}{c|}{Sun360 Outdoor dataset} \\
\cline{2-9}
            & PSNR & SSIM &FSIM &Q score  & PSNR & SSIM &FSIM &Q score  \\
\hline
Hdrcnn~\cite{Eilertsen-tog17-hdrcnn}   &11.99    &0.2249    &0.5687     &39.64         &11.09    &0.6007     &0.8637     &56.31        \\
DrTMo~\cite{Endo-tog17-DrTMO}          &-        &-         &-          &-             &14.64    &0.6822     &0.8101     &52.39        \\
\rowcolor[gray]{.7} DRHT                                   &28.18    &0.9242    &0.9622     &97.87         &22.60    &0.7629     &0.8691     &56.17        \\
\hline
\end{tabular}
\end{center}
\vspace{-3.0mm}
\caption{Quantitative evaluation between the proposed DRHT method and two HDR prediction methods. The results of DrTMo on the City Scene Dataset are not available as it requires high resolution inputs. The evaluation indicates the proposed DRHT method is effective to generate HDR data compared with existing HDR prediction methods.}
\label{tab:extensive}
\end{table*}

\subsection{Comparison with State-of-the-arts}
We compare the proposed DRHT method with state-of-the-art image correction methods on the standard benchmarks. The visual evaluation is shown in Figure~\ref{fig:overCompare} where the input images are captured in over exposure. The image filtering based methods are effective to preserve local edges. However, they cannot recover the details in the overexposed regions, as shown in (c), (d) and (f). It is because these methods tend to smooth the flat region while preserving the color contrast around the edge region. They fail to recover the details, which reside in the overexposed regions where the pixel values approach 255. Meanwhile, the image correction methods based on global contrast and saturation manipulation are not effective as shown in (r). They share the similar limitations as image filtering based methods as the pixel-level operation fails to handle overexposed images. The results of WVM~\cite{Fu-CVPR16-WVM} tend to be brighter as shown in (e), (m) and (u) as they over enhance the illumination layer decomposed from the input image. Compared with existing methods, the proposed DRHT method can successfully recover the missing details buried in the over exposed regions while maintaining the realistic global illumination.

Figure~\ref{fig:underCompare} shows some under/over exposed examples in the low-light scenes. It shows that the image filtering based methods can only strengthen existing details. CAPE~\cite{Kaufman-cgf12-CAPE} performs well in the low-light regions as shown in (b) but it simply adjusts the brightness and thus fails to correct all missing details. Figure~\ref{fig:underCompare}(i) shows that WVM~\cite{Fu-CVPR16-WVM} performs poorly in the scenes with dark skies, as it fails to decompose the dark sky into reflectance and the illumination layers. Meanwhile, the missing details in the under/over exposed regions can be reconstructed via the proposed DRHT method as shown in (h) and (p). Global illumination is also maintained through residual learning.

We note that the proposed DRHT method tends to slightly increase the intensity in the dark regions. There are two reasons for this. First, DRHT is trained on the city scene dataset~\cite{Zhang-ICCV17-ldr2hdr}, where the sun is always located near the center of the images. Hence, when the input image has some bright spots near to the center, the night sky will tend to appear brighter as shown in Figure~\ref{fig:underCompare}(p)). Second, as we use the first network to predict the gamma compressed HDR image and then map it back to the LDR in the logarithmic domain, low intensity values may be increased through the inverse gamma mapping and logarithmic compression as shown in Figure~\ref{fig:underCompare}(h).

In additional to visual evaluation, we also provide quantitative \ryn{comparison} between the proposed method and existing methods as summarized in Table~\ref{tab:psnr}. It shows that the proposed method performs favorably against existing methods under several numerical evaluation metrics.

\def\swfive{0.2\linewidth}
\begin{figure*}
	\begin{center}
    \begin{tabular}{ccccc}
			\includegraphics[width=\swfive]{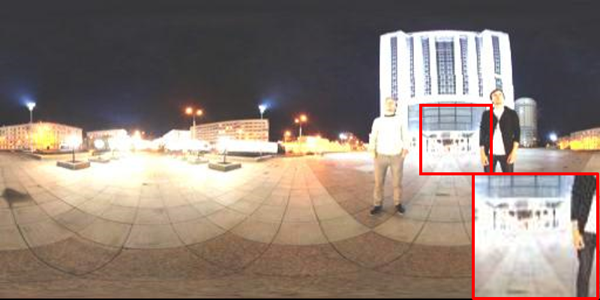}&
			\includegraphics[width=\swfive]{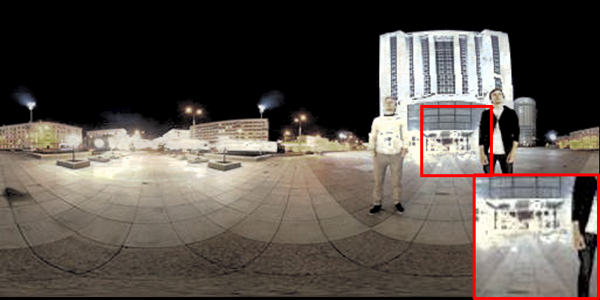}&
			\includegraphics[width=\swfive]{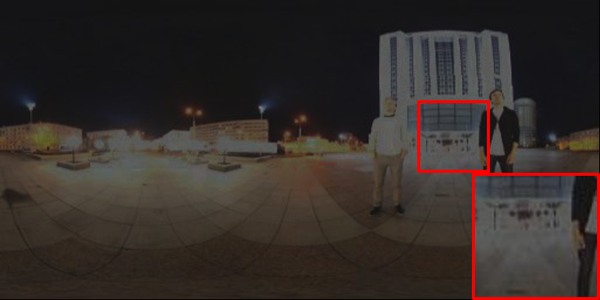}&
			\includegraphics[width=\swfive]{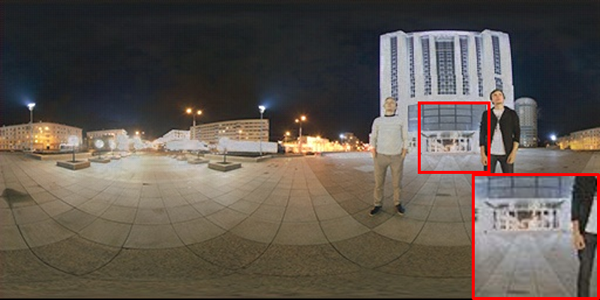}&
			\includegraphics[width=\swfive]{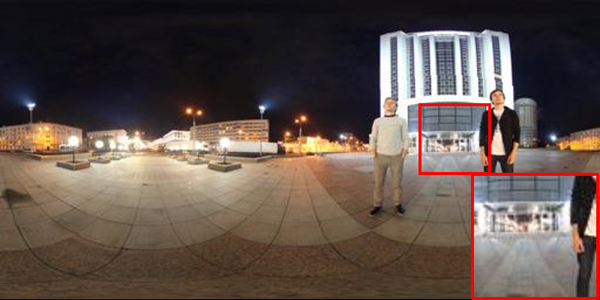}\\
            (a) Input  &(b) DrTMo~\cite{Endo-tog17-DrTMO} &(c) Hdrcnn~\cite{Eilertsen-tog17-hdrcnn} &(d) DRHT &(e) Ground Truth\\
			\includegraphics[width=\swfive]{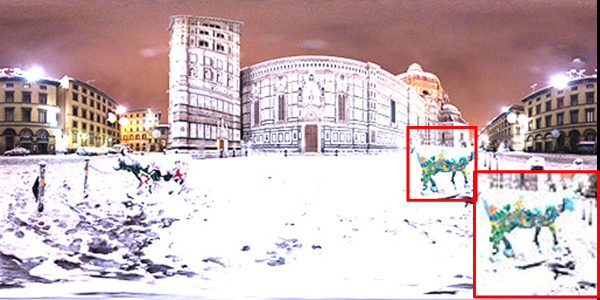}&
			\includegraphics[width=\swfive]{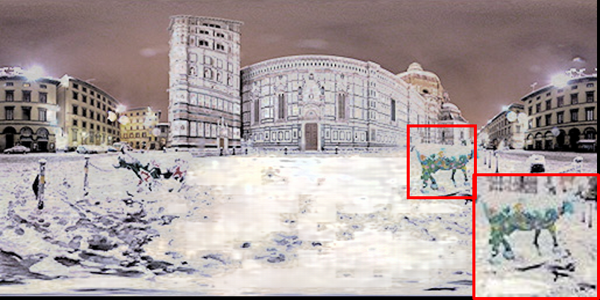}&
			\includegraphics[width=\swfive]{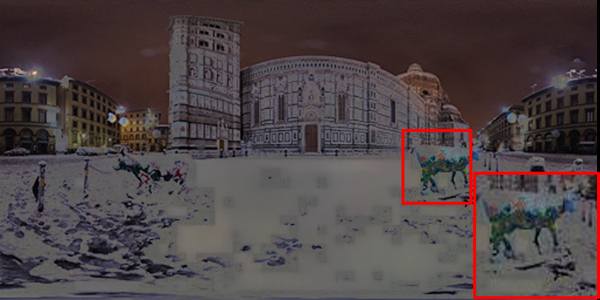}&
			\includegraphics[width=\swfive]{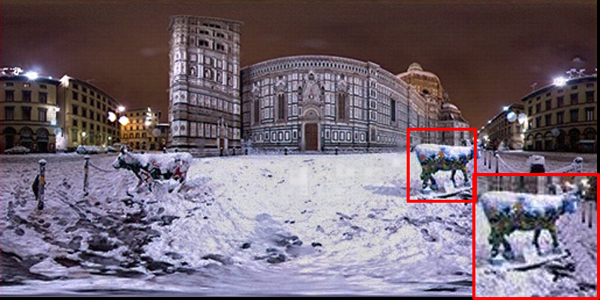}&
            \includegraphics[width=\swfive]{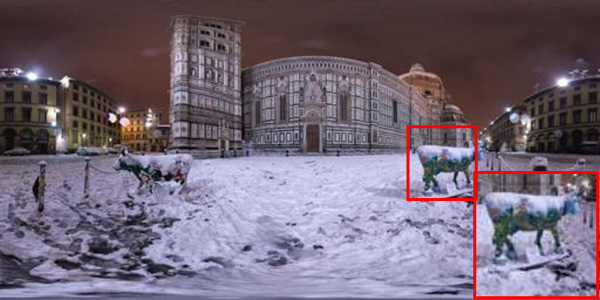}\\
            (e) Input  &(f) DrTMo~\cite{Endo-tog17-DrTMO} &(g) Hdrcnn~\cite{Eilertsen-tog17-hdrcnn} &(h) DRHT &(e) Ground Truth\\
    \end{tabular}
    \end{center}
    \vspace{-3mm}
    \caption{Visual \ryn{comparison} with two HDR based correction methods: DrTMo~\cite{Endo-tog17-DrTMO} and Hdrcnn~\cite{Eilertsen-tog17-hdrcnn}, on the Sun360 outdoor dataset. The proposed DRHT performs better than these two methods in generating visually pleasing images. 
    }
	\label{fig:DrTMO}
\end{figure*}

We further compare the proposed DRHT method with two HDR prediction methods (\ie, DrTMO~\cite{Endo-tog17-DrTMO} and Hdrcnn~\cite{Eilertsen-tog17-hdrcnn}). These two methods can be treated as image correction methods because their output HDR image can be tone mapped into the LDR image. In~\cite{Endo-tog17-DrTMO}, two deep networks are proposed to first generate up-exposure and down-exposure LDR images from the single input LDR image. As each image with limited exposure cannot contain all the details of the scene to solve the under/over exposure problem, they fuse these multiple exposed images and use \cite{Kim-CGIM08-CTR} to generate the final LDR images. Eilertsen~\etal~\cite{Eilertsen-tog17-hdrcnn} propose a deep network to blend the input LDR image with the reconstructed HDR information in order to recover the high dynamic range in the LDR output images. However, by using the highlight masks for blending, their method cannot deal with the under exposed regions and their results tend to be dim as shown in Figures~\ref{fig:DrTMO}(c) and~\ref{fig:DrTMO}(g). Meanwhile, we can also observe obvious flaws in the output images of both DrTMO~\cite{Endo-tog17-DrTMO} and Hdrcnn~\cite{Eilertsen-tog17-hdrcnn} (e.g., the man's white shirt in Figure~\ref{fig:DrTMO}(b) and the blocking effect in the snow in Figure~\ref{fig:DrTMO}(g)). The main reason lies in that existing tone mapping methods fail to preserve the local details from the HDR domain when the under/exposure exposure problem happens. In comparison, the proposed DRHT is effective to prevent this limitation because we do not attempt to recover the whole HDR image but only focus on recovering the missing details by residual learning. The quantitative evaluation results shown in Table~\ref{tab:extensive} indicate that the proposed DRHT method performs favorably against these HDR prediction methods.

\subsection{Limitation Analysis}
Despite the aforementioned success, the proposed DRHT method contains limitation to recover the details when significant illumination contrast appears on the input images. Figure~\ref{fig:fail} shows one example. Although DRHT can effectively recover the missing details of the hut in the underexposed region (i.e., the red box in Figure~\ref{fig:fail}), there are limited details around the sun (i.e., the black box). This is mainly because of the large area of overexposed sunshine is rare in our training dataset. In the future, we will augment our training dataset to incorporate such extreme cases to improve the performance.

\def\swtwo{0.5\linewidth}
\begin{figure}
	\begin{center}
    \begin{tabular}{cc}
			\includegraphics[width=\swtwo,height=0.8in]{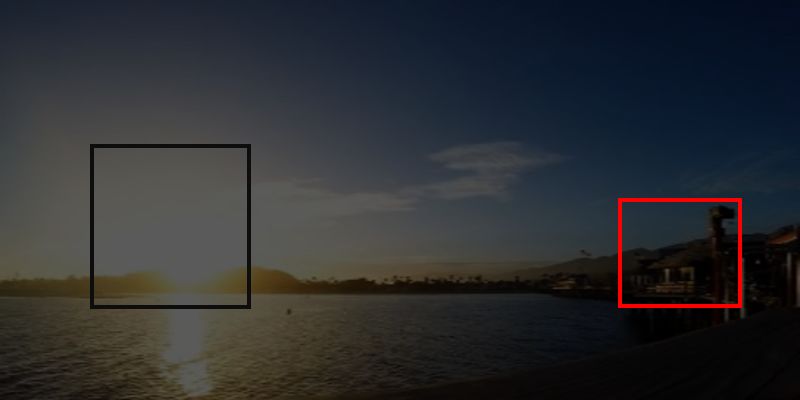}&
			\includegraphics[width=\swtwo,height=0.8in]{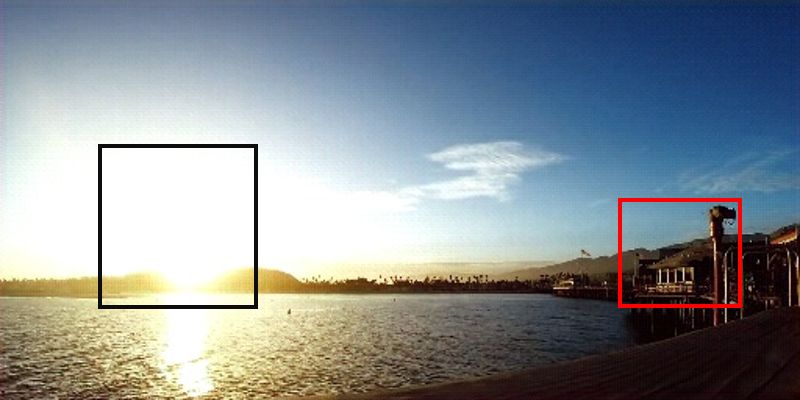}\\
            (a) Input  &(b) DRHT\\
    \end{tabular}
    \end{center}
    \vspace{-3mm}
    \caption{Limitation analysis. The proposed DRHT method is effective to recover the missing details in the underexposed region marked in the red box, while limits on the overexposed sunshine region marked in the black box.}
	\label{fig:fail}
\end{figure}

\section{Conclusion}
In this paper, we propose a novel deep reciprocating HDR transformation (DRHT) model for under/over exposed image correction. We first trace back to the image formulation process to explain why the under/over exposure problem is observed in the LDR images, according to which we reformulate the image correction as the HDR mapping problem. We show that the buried details in the under/over exposed regions cannot be completely recovered in the LDR domain by existing image correction methods. Instead, the proposed DRHT method first revisits the HDR domain and recovers the missing details of natural scenes via the HDR estimation network, and then transfers the reconstructed HDR information back to the LDR domain to correct the image via another proposed LDR correction network. These two networks are formulated in an end-to-end manner as DRHT and achieve state-of-the-art correction performance on two benchmarks.

\section*{Acknowledgements}
We thank the anonymous reviewers for the insightful and constructive comments, and NVIDIA for generous donation of GPU cards for our experiments. This work is in part supported by an SRG grant from City University of Hong Kong (Ref. 7004889), and by NSFC grant from National Natural Science Foundation of China (Ref. 91748104, 61632006, 61425002).

\newpage

{\small
\bibliographystyle{ieee}
\bibliography{ref}
}

\end{document}